\documentclass{article}

\PassOptionsToPackage{numbers}{natbib}

\usepackage[preprint]{neurips_2026}

\usepackage[utf8]{inputenc} 
\usepackage[T1]{fontenc}    
\usepackage{hyperref}       
\usepackage{url}            
\usepackage{booktabs}       
\usepackage{amsfonts}       
\usepackage{nicefrac}       
\usepackage{microtype}      
\usepackage{xcolor}         
\usepackage{microtype}
\usepackage{graphicx}
\usepackage{subcaption}
\usepackage{booktabs} 
\usepackage{makecell}
\usepackage{listings}
\usepackage{enumitem}
\usepackage{amsmath}

\usepackage{tabularx}
\usepackage{booktabs}
\usepackage{array}
\title{Agentic AI Systems Should Be Designed as Marginal Token Allocators}

\author{%
  Siqi Zhu \\
  University of Illinois Urbana-Champaign \\
  \texttt{siqizhu4@illinois.edu} \\
}

\begin{document}

\maketitle

\begin{abstract}
This position paper argues that agentic AI systems should be designed and evaluated as \emph{marginal token allocation economies} rather than as text generators priced by the unit. We follow a single request---a developer asking a coding agent to fix a failing test---through four economic layers that today are designed in isolation: a router that decides which model answers, an agent that decides whether to plan, act, verify, or defer, a serving stack that decides how to produce each token, and a training pipeline that decides whether the trace is worth learning from. We show that all four layers are solving the \emph{same} first-order condition---marginal benefit equals marginal cost plus latency cost plus risk cost---with different index sets and different prices. The framing is deliberately minimal: we do not propose a complete theory of AI economics. But adopting marginal token allocation as the shared accounting object explains why systems that locally minimize tokens globally misallocate them, predicts a small set of recurring failure modes (over-routing, over-delegation, under-verification, serving congestion, stale rollouts, cache misuse), and points to a concrete research agenda in token-aware evaluation, autonomy pricing, congestion-priced serving, and risk-adjusted RL budgeting.
\end{abstract}

\section{Introduction}
\label{sec:intro}

Consider a developer who types ``the CI test on \texttt{auth/login} is failing---fix it'' into a modern coding agent. Before a single line of code is touched, the system has already made four economic decisions. A \emph{router} decides whether to spend a cheap model (fast triage, possibly wrong) or a frontier model (slow, expensive, more likely correct) \citep{chen2023frugalgpt,ong2024routellm}. An \emph{agent policy} decides how the chosen model should spend its tokens---reading the repository, planning, editing, running tests, or asking the developer to clarify \citep{yao2023react,shinn2023reflexion,wang2024voyager}. A \emph{serving stack} decides how to produce those tokens, juggling prefill for the long context, decode for the patch, and KV cache for the test logs \citep{kwon2023efficient,patel2024splitwise,zhong2024distserve,fu2024efficientllmschedulinglearning}. And a \emph{training pipeline} decides, after the dust settles, whether this trace is worth learning from---rollout, verifier, or update tokens to spend now for capability later \citep{ouyang2022training,deepseek2025r1,chang2026srtacceleratingreinforcementlearning,zhu2026opentinkerseparatingconcernsagentic}.

Each layer charges a different price for what looks, on the API invoice, like the same token. The router prices a token in dollars per million; the agent prices it in expected risk of an irreversible action; the serving stack prices it in queueing delay; the trainer prices it in marginal capability gain over a discount horizon. This decoupling is hidden by the dominant accounting fiction---tokens are units of text, billed at a flat rate \citep{brown2020language}. That fiction was workable when LLMs were chat completions. It is misleading once tokens cause actions, occupy infrastructure, and become training data.

This position paper argues that \emph{agentic AI systems should be designed and evaluated as marginal token allocation economies}, in which routers, agents, serving schedulers, and trainers are mechanisms that decide where the next unit of tokenized computation should be spent under joint quality, cost, latency, and risk constraints. The claim is narrower than ``token economics is a complete theory of AI'' and stronger than ``tokens are billed by the unit.'' We argue that a \emph{single} first-order condition---marginal benefit equals marginal cost plus latency cost plus risk cost---is the right minimum vocabulary, because the four layers above are not parallel engineering problems but vertical slices of one allocation problem. The router screens the demand side, the agent contracts on the action side, the serving stack produces on the supply side, and the trainer accumulates capital on the investment side. They are the same equation, evaluated at four shadow prices that today no single layer can see.

\paragraph{The central tension.}
Each layer optimizes locally and competently. Routers minimize cost subject to quality \citep{chen2023frugalgpt}; agents maximize success rate \citep{liu2024agentbench}; serving stacks maximize throughput \citep{agrawal2024sarathi}; trainers maximize evaluation score \citep{deepseek2025r1}. Yet local rationality aggregates into global misallocation: an aggressive router downgrades a high-stakes request, the agent compensates by burning extra verification tokens, the serving stack queues those verifier calls behind unrelated long-context traffic, and the trainer learns from a noisy trace that will not generalize. The pattern is the textbook problem of unpriced externalities \citep{pigou1920economics,coase1937nature}, transposed to token economies. Marginal token allocation is the shared price language that lets the four layers cooperate rather than merely stack.

\paragraph{Contributions.}
(i) We formulate a single optimality condition---marginal token allocation---and show that routers, agents, serving stacks, and trainers are instances of it (Section~\ref{sec:theory}). (ii) We trace one request through all four layers, using standard tools from microeconomics: screening with hidden types \citep{akerlof1970market,spence1973job}, principal--agent contracts \citep{mirrlees1976optimal,holmstrom1979moral}, multi-stage production with congestion \citep{pigou1920economics,vickrey1969congestion}, and capital accumulation \citep{solow1956contribution} (Section~\ref{sec:layers}). (iii) We show that recurring failures across the stack---over-routing, over-delegation, under-verification, congestion, stale rollouts, cache misuse---are corner cases of the same equation when one of the four prices is mis-set (Section~\ref{sec:failures}). (iv) We address principled objections (Section~\ref{sec:alt}), discuss design implications, limitations, and an open research agenda (Section~\ref{sec:discussion}), and conclude (Section~\ref{sec:conclusion}). Throughout, our objective is not to summarize the literature but to defend a single design stance.

\section{One Equation, Four Prices}
\label{sec:theory}

\paragraph{The primitive object.}
Let an LLM \emph{system} face a stream of tasks. For each task it has a finite set of \emph{token uses}, indexed by $i$, between which it must allocate computation. Concretely, $i$ ranges over choices such as \{cheap model, frontier model, retrieval, planning, tool call, verifier, prefill capacity, decode capacity, KV transfer, RL rollout, reward computation, gradient update\}. Each use $i$ has a marginal quality contribution $\Delta Q_i$, a marginal compute cost $\Delta C_i$, a marginal latency cost $\Delta L_i$, and a marginal risk $\Delta R_i$ (e.g., probability of a wrong action weighted by its consequence). Let $V$ denote task value. The system should spend the next token on
\begin{equation}
\label{eq:master}
    i^{*} \;=\; \arg\max_{i}\;\Big[\, V\,\Delta Q_i \;-\; \Delta C_i \;-\; \lambda\,\Delta L_i \;-\; \rho\,\Delta R_i\,\Big],
\end{equation}
where $\lambda \ge 0$ and $\rho \ge 0$ are user- or operator-specific shadow prices on latency and risk. Equation~\ref{eq:master} is the standard marginal-utility decision rule of microeconomics \citep{mascolell1995micro} transposed to tokenized computation. At an interior optimum, the Marshallian equimarginal condition holds:
\begin{equation}
\label{eq:equim}
    V\,\Delta Q_i \;-\; \lambda\,\Delta L_i \;-\; \rho\,\Delta R_i \;=\; \Delta C_i \quad \forall\, i \in \mathcal{A}^{*},
\end{equation}
where $\mathcal{A}^{*}$ is the set of token uses with strictly positive allocation. ``The marginal benefit of a token equals its full marginal cost,'' once latency and risk are properly priced.

\paragraph{Why the four prices live at four layers.}
Equation~\ref{eq:master} packs the entire stack into one expression, but its terms are observed at different layers (Table~\ref{tab:prices}). $V$ is set by the user, who alone knows the value of her task; $\Delta C_i$ is set by the operator, who runs the GPUs; $\lambda$ is set by the SLA, which arbitrates queueing; $\rho$ is set by the safety team, which absorbs the consequences of wrong actions. No single layer sees all four. This is the structural reason why locally rational decisions compose into globally irrational allocations \citep{tirole1988theory}, and why a shared accounting object is needed.

\begin{table}[t]
  \caption{The same allocation primitive is observed at four organizational layers, each of which sees only one or two of the four prices in Equation~\ref{eq:master}. Marginal token allocation is the language that makes the four layers commensurable.}
  \label{tab:prices}
  \centering
  \small
  \setlength{\tabcolsep}{6pt}
  \begin{tabular}{@{}lllll@{}}
    \toprule
    Layer & Mechanism & Index $i$ & Price observed &  Paragraph \\
    \midrule
    Demand   & Routing as screening    & model tier         & $V$, $\Delta C_i$    & \S\ref{sec:demand} \\
    Action   & Agent as principal--agent & plan/act/verify  & $\rho$, $V$           & \S\ref{sec:action} \\
    Supply   & Serving as production   & prefill/decode/KV  & $\lambda$, $\Delta C_i$ & \S\ref{sec:supply} \\
    Capital  & Caches \& RL as investment & rollout/store    & $\Delta C_i$, $\rho$   & \S\ref{sec:capital} \\
    \bottomrule
  \end{tabular}
\end{table}

\paragraph{Why ``marginal'' rather than ``total''.}
Industry dashboards typically report total or average token cost. But a 30\% reduction in total tokens that comes from cutting verifier tokens may \emph{raise} risk-adjusted cost, because the cost of an unverified wrong action exceeds the savings. Marginal analysis makes this explicit: the right object is $\partial U / \partial t_i$, not $U / \sum_i t_i$. We will see this gap is precisely where current systems misallocate.

\paragraph{A worked example.}
A small numerical instance of Equation~\ref{eq:master} clarifies the stakes. Suppose two models are available: a cheap one with quality $q_c = 0.7$ at cost $c_c = 1$, and a frontier one with $q_f = 0.9$ at $c_f = 5$. For a low-value task ($V = 10$), surplus is $0.7 \cdot 10 - 1 = 6$ versus $0.9 \cdot 10 - 5 = 4$, so cheap wins. For a high-value task ($V = 100$), surpluses are $69$ versus $85$, and frontier wins. The crossover is at $V^{*} = (c_f - c_c)/(q_f - q_c) = 20$. Now add risk: if the cheap model has a wrong-action probability $r_c = 0.05$ versus $r_f = 0.01$ and risk price $\rho = 50$, the cheap-versus-frontier surplus gap shrinks by $\rho(r_c-r_f) = 50 \cdot 0.04 = 2$, shifting $V^{*}$ to $\approx 10$. A small change in one term in Equation~\ref{eq:master} flips the optimal allocation. This is why marginal analysis is non-trivial in practice: each layer adjusts a different term, and small shifts compound across layers.

\paragraph{A closed-form sanity check.}
A Cobb--Douglas instance $Q(\mathbf{t}) = A\,\prod_i t_i^{\alpha_i}$ subject to $\sum_i p_i t_i \le B$, with full shadow price $p_i = \Delta C_i + \lambda \Delta L_i + \rho \Delta R_i$, has the textbook solution $t_i^{*} = \tfrac{\alpha_i}{\sum_j \alpha_j}\cdot \tfrac{B}{p_i}$ \citep{mascolell1995micro}. Three operational facts follow: irrelevant uses ($\alpha_i=0$) consume zero tokens regardless of price; rising $p_i$ proportionally squeezes use $i$, the substitution pattern observed in production schedulers \citep{agrawal2024sarathi}; and complements cannot be cut to zero without driving $Q$ to zero---which is why ``minimize tokens'' fails when reading and verification complement editing.

\paragraph{Prices as Lagrange multipliers, and the welfare-theorem prescription.}
The four prices in Equation~\ref{eq:master} are not chosen by fiat; they are the dual variables of the constrained primal of token allocation. Consider a system maximizing $\sum_x V(x)\,Q(\mathbf{t}_x)$ subject to a compute budget, a latency SLA, and a risk envelope. The Lagrangian
\begin{equation}
\label{eq:lagrangian}
\mathcal{L} = \sum_{x} V(x)\,Q(\mathbf{t}_x) - \mu_C \!\left(\!\sum_{x,i}\! \Delta C_i\,t_{i,x} - \bar{C}\right) - \mu_L \!\left(\!\sum_{x,i}\! \Delta L_i\,t_{i,x} - \bar{L}\right) - \mu_R \!\left(\!\sum_{x,i}\! \Delta R_i\,t_{i,x} - \bar{R}\right)
\end{equation}
yields KKT stationarity $V(x)\,\partial Q/\partial t_{i,x} = \mu_C \Delta C_i + \mu_L \Delta L_i + \mu_R \Delta R_i$ at the optimum, which is exactly Equation~\ref{eq:master} with $(1, \lambda, \rho) = (\mu_C, \mu_L/\mu_C, \mu_R/\mu_C)$. Three implications follow. First, the prices are \emph{endogenous}: determined by the binding constraints, not chosen a priori. Second, they obey complementary slackness, so a system whose latency SLA slacks should drive $\lambda \to 0$ rather than pin it to a constant---the standard production-stack practice. Third, by the first welfare theorem \citep{mascolell1995micro}, if router, agent, serving stack, and trainer all maximize their own component of $\mathcal{L}$ taking the same $(\mu_C, \mu_L, \mu_R)$ as given, the resulting allocation is Pareto efficient: no reallocation of tokens across layers improves any payoff without hurting another's. The second welfare theorem implies that any efficient allocation can be sustained by some price vector. Together they yield a sharp design prescription: the question is not whether to centralize allocation but whether the four layers see a common, complete price vector. They almost never do today.

\paragraph{Information rents and the screening cost of routing.}
The router is not just choosing; it is screening. A type-$\theta$ user knows her own $V$ but the router does not. Mechanism design \citep{mirrlees1976optimal,laffont2002theory} then implies that the cost of truthful self-selection is an \emph{information rent} paid to high-value types: with type distribution $F(\theta)$ and the increasing-hazard property, the optimal menu prices the marginal type at virtual valuation $V(\theta) - \tfrac{1-F(\theta)}{f(\theta)}\,V'(\theta)$ rather than at $V(\theta)$. The wedge $\tfrac{1-F}{f}\,V'$ does not appear in any naive cost--quality dashboard. Two empirical implications follow. Even an optimally designed router will downgrade a non-trivial fraction of high-$V$ requests on purpose: the rent is the price of incentive compatibility, not a bug. And the rent grows in user heterogeneity, which is why a router tuned on a uniform benchmark systematically fails on long-tail traffic.

\paragraph{General equilibrium across tenants.}
A single request faces four prices, but multi-tenant deployments must clear them simultaneously. A competitive equilibrium \citep{mascolell1995micro,tirole1988theory} is a price vector $\mathbf{p}^{*}$ such that each tenant's demand $\mathbf{z}_x(\mathbf{p}^{*})$ solves its own Equation~\ref{eq:master}, the operator's supply maximizes profit given the production frontier, and markets clear, $\sum_x \mathbf{z}_x(\mathbf{p}^{*}) = \mathbf{Y}(\mathbf{p}^{*})$. The first welfare theorem then guarantees that the equilibrium allocation is Pareto efficient \emph{across tenants}, internalizing the queueing externality that flat per-token pricing cannot. The closest production analogues are priority queues with admission control \citep{agrawal2024sarathi,fu2024efficientllmschedulinglearning}, equivalent to a degenerate equilibrium in which only one constraint is priced.

\paragraph{The Knightian limit of $\rho$.}
A separate caveat applies to $\rho \Delta R_i$ itself: it captures expected-value risk, but agentic actions are often novel and their consequence distribution unknown---a Knightian regime \citep{knight1921risk}. The framework is not changed, but the functional form of $\rho \Delta R_i$ should switch from expectation to a coherent risk measure (e.g., CVaR or max-min over an ambiguity set) on rare, high-consequence actions.

\paragraph{What the theory does \emph{not} claim.}
We deliberately stop at a first-order condition. Equation~\ref{eq:master} is a one-step rule; we do not claim that summing it across tasks gives a complete macroeconomic theory of AI, nor that token allocation is the only relevant economic primitive (data, energy, and labor matter too). We use marginalism as a \emph{lens}: it should produce sharp predictions for system design and identify shared structure across what otherwise look like unrelated engineering problems.

\section{One Request, Four Layers}
\label{sec:layers}

We now follow the developer's request from \S\ref{sec:intro} through each layer of the stack and show that each layer is solving Equation~\ref{eq:master} at a different price. The narrative deliberately preserves the request's identity: the same task picks up new prices as it descends.

\subsection{Demand: Routing as a Screening Mechanism}
\label{sec:demand}

The first decision is which model answers. Naively one would route by ``best quality per dollar.'' That intuition is wrong in the same way that posting a single price is wrong in a market with heterogeneous buyers \citep{tirole1988theory}.

A request has a hidden type $\theta = (V, d, r, \lambda)$: task value, difficulty, risk sensitivity, latency sensitivity. The router observes only $x$, a noisy signal of $\theta$. Its problem is the screening problem of \citet{spence1973job} and \citet{mirrlees1976optimal}: design a mapping $m^{*}(x)$ such that the chosen model maximizes Equation~\ref{eq:master} restricted to the model index,
\begin{equation}
\label{eq:routing}
    m^{*}(x) \;=\; \arg\max_{m \in \mathcal{M}} \;\Big[\, V(x)\, \widehat{q}_m(x) \;-\; c_m \;-\; \lambda\, l_m(x) \;-\; \rho\, r_m(x)\,\Big].
\end{equation}
Recent routers \citep{chen2023frugalgpt,ong2024routellm,hu2024routerbench} estimate $\widehat{q}_m(x)$ from preference data or cascades. That estimation is doing economic work: it converts a flat ``model market'' into a differentiated market in which each request is matched to the cheapest model that preserves expected utility. \citet{akerlof1970market} showed that hidden quality on the seller side can collapse a market; routing exposes the symmetric problem on the buyer side, where hidden \emph{difficulty} causes mis-matched models. Both directions are observed in production \citep{ong2024routellm,hu2024routerbench}.

In our running example, the router must guess whether the failing-test query is shallow (cheap-model territory) or deep (frontier territory) from a few hundred characters of prompt. If it guesses shallow and the bug is a subtle race condition, the agent will burn tokens later trying to compensate, the developer will eventually re-issue the request to a stronger model, and the system will pay for both attempts. If it guesses deep and the bug is a forgotten import, the operator overpays by a factor of five (using our worked numbers from \S\ref{sec:theory}). The cost of routing error is therefore not symmetric: the misallocation propagates downstream and is amortized by every layer below. Strategic users know this, and a sophisticated developer can perturb $x$ to obtain a stronger model---an LLM analogue of Spence's costly signaling. A revenue-equivalent design would charge a premium for higher tiers and let users self-select via an incentive-compatible menu \citep{tirole1988theory},
\begin{equation}
\label{eq:ic}
    V_k\, q_{m_k} - p_k \;\ge\; V_k\, q_{m_{k'}} - p_{k'}, \qquad \forall\, k, k'.
\end{equation}
Few production routers do this; most attempt to infer $\theta$ silently.

\paragraph{Position.}
Routers should be evaluated not by accuracy or cost alone but by \emph{regret} relative to Equation~\ref{eq:routing}: the gap between the chosen model's risk-adjusted utility and the ex-post optimal model. They should publish either the regret bound or the menu (Eq.~\ref{eq:ic}). Existing benchmarks \citep{hu2024routerbench} approximate the former only loosely and almost never report risk components.

\subsection{Action: Agents as Principal--Agent Contracts}
\label{sec:action}

The chosen model now enters the agent loop. The router has answered ``which model''; the agent must answer ``what should it do.'' This is where the same token's price changes again, because tokens used to summarize a file and tokens used to commit a patch carry different consequences \citep{yao2023react,schick2023toolformer,wang2024voyager}.

\paragraph{The autonomy contract.}
Let $a \in [0,1]$ denote autonomy (0 = always ask, 1 = act freely) and $t$ be the token budget. The user's expected utility is
\begin{equation}
\label{eq:agent}
    U(a, t) \;=\; V\, p(a, t) \;-\; C(t) \;-\; R(a, t) \;-\; H(a),
\end{equation}
where $p(a, t)$ is success probability, $C(t)$ is the token cost, $R(a, t)$ is the expected loss from autonomous mistakes, and $H(a)$ is the human-oversight cost (decreasing in $a$). The interior optimality condition is the principal--agent first-order condition \citep{mirrlees1976optimal,holmstrom1979moral,laffont2002theory}:
\begin{equation}
\label{eq:autonomy}
    V\, \frac{\partial p}{\partial a} \;=\; \frac{\partial R}{\partial a} \;+\; \frac{\partial H}{\partial a}.
\end{equation}
Autonomy expands until the marginal value of saved human labor equals the marginal increase in risk plus the marginal change in oversight cost. Because $\partial R/\partial a$ is heavily right-skewed---small probability of a catastrophic action---risk-neutral budgeting badly under-prices autonomy.

\paragraph{Token allocation \emph{within} the agent.}
Once $a$ is set, the agent still has a team-production problem \citep{alchian1972production}. In our example, the agent must split tokens among reading the repo ($T_r$), planning the patch ($T_p$), editing ($T_e$), and running the test ($T_v$):
\begin{equation}
    Y \;=\; F(T_r, T_p, T_e, T_v, H_{\text{review}}).
\end{equation}
At the optimum, marginal products are equalized: $\partial Y/\partial T_r = \partial Y/\partial T_p = \partial Y/\partial T_e = \partial Y/\partial T_v$. This contradicts the heuristic of ``minimize tokens.'' Reading and verification tokens are \emph{complements} to editing tokens \citep{shinn2023reflexion,madaan2023self,lightman2024lets,fu2025efficientlyscalingllmreasoning}: the marginal product of an edit token is small without context and verification, and the marginal product of additional reasoning tokens is itself task-dependent---a fact that signals such as model certainty \citep{fu2025efficientlyscalingllmreasoning} can be used to estimate online. Empirically, agents that skimp on $T_v$ produce cheaper but lower-quality patches and shift cost downstream onto $H_{\text{review}}$. The team-production view also explains why imitating only the editing step from a strong model rarely transfers: the chain of complements upstream and downstream is what produces $Y$, and a partial copy is not Pareto-improving.

\paragraph{Reversibility and option value.}
Because action risk is partly irreversible, autonomy decisions also carry option value, in the sense of the real-options literature \citep{dixit1994investment}. Asking the user for confirmation preserves the option to act later; acting immediately destroys it. Equation~\ref{eq:master} should therefore include an additional term $\rho_{\text{irr}} \Delta R^{\text{irr}}_i$ for the unrecoverable component of risk. This is why ``read'' and ``draft'' tokens flow freely while ``commit'' and ``send'' deserve a discrete oversight check.

\paragraph{Position.}
Agentic systems should publish an \emph{autonomy schedule}---a mapping from action class to required confirmation level (read $\to$ free, draft $\to$ free, commit $\to$ confirm, deploy/transfer $\to$ multi-party). It is the LLM equivalent of an authorization matrix and is the natural artifact of Equation~\ref{eq:autonomy}. Current agentic benchmarks \citep{liu2024agentbench} measure success rate but rarely measure $R(a, t)$, which we argue is the binding economic constraint.

\subsection{Supply: Serving as Production}
\label{sec:supply}

Each token the agent commands must be physically produced. Modern stacks separate prefill and decode \citep{patel2024splitwise,zhong2024distserve}, page KV cache \citep{kwon2023efficient}, and chunk requests \citep{agrawal2024sarathi}. Speculative decoding \citep{leviathan2023fast} adds a verifier stage. These are exactly the moves a microeconomic theorist would predict in a multi-stage production system with heterogeneous resources \citep{alchian1972production}.

Let $G_p, G_d, K, N$ denote prefill GPU capacity, decode GPU capacity, KV-cache storage/bandwidth, and interconnect bandwidth. Token output is $Y_{\text{tok}} = F(G_p, G_d, K, N)$ and latency $L = L_p(G_p) + L_d(G_d) + L_K(K, N)$. The cost-minimizing producer satisfies the equimarginal condition,
\begin{equation}
\label{eq:serving}
    \frac{\partial L / \partial G_p}{\partial C / \partial G_p}
    \;=\;
    \frac{\partial L / \partial G_d}{\partial C / \partial G_d}
    \;=\;
    \frac{\partial L / \partial K}{\partial C / \partial K},
\end{equation}
i.e., latency reduction per dollar should be equalized across resources. \citet{patel2024splitwise} and \citet{zhong2024distserve} show empirically that pre-disaggregation systems were systematically off this frontier.

In our example, the agent's plan-then-edit-then-test loop stresses the supply layer in characteristic patterns. Reading the repo is prefill-heavy. Generating the patch is decode-heavy. Running the test produces a long error log---prefill again. None of these is the same token, economically. A request that occupies a long-context KV cache imposes a queueing externality on every other tenant---a textbook congestion externality \citep{pigou1920economics,vickrey1969congestion}. The first-best policy is congestion pricing: charge each request the marginal external delay it imposes. Most production APIs charge a flat per-token rate, which under-prices long-context, decode-heavy traffic and over-prices short prompts. Recent schedulers that learn to rank requests by predicted output length \citep{fu2024efficientllmschedulinglearning} are an early step in this direction: they transform an unpriced FCFS queue into something closer to a priority discipline that internalizes the queueing externality, even if the implied prices are not surfaced to the upstream router or agent.

The serving layer also reveals why the previous two layers' decisions cannot be evaluated in isolation. The router that selected the frontier model at the demand layer has implicitly committed the supply layer to higher prefill cost; the agent that chose ``read the whole repo before planning'' has implicitly committed it to higher KV-cache pressure. If the supply layer's prices $\Delta C_i$ are not visible upstream, the demand and action layers will optimize as if compute were free, and the supply layer will absorb the externality as queueing delay. This is the operational mechanism by which \emph{one} layer's local optimum becomes \emph{another} layer's congestion problem.

\paragraph{Speculative decoding as outsourced labor.}
Speculative decoding is a make-or-buy decision \citep{coase1937nature}: a cheap draft model produces candidate tokens that the expensive model verifies. The arrangement is profitable when the verifier's marginal cost of accepting a draft is strictly less than its marginal cost of generating from scratch. The acceptance rate $\alpha$ plays the role of an internal transfer price; small drops in $\alpha$ flip the make-or-buy calculus. Adversarially long contexts---where $\alpha$ falls---should disable speculation rather than merely slow it. This is the textbook Coasean prediction: integration dominates the market when transaction costs are high.

\paragraph{Position.}
Serving systems should expose, log, and ideally bill against \emph{shadow prices} for prefill, decode, and KV resources. These shadow prices are the operational manifestation of $\Delta C_i$ in Equation~\ref{eq:master} and are a prerequisite for upstream layers (the router and the agent) to make correct decisions.

\subsection{Capital: Caches and RL Training as Investment}
\label{sec:capital}

After the developer's test passes, two streams of tokens persist. The KV blocks for the repo prefix and the test logs may be cached for the next request; the trace itself may be added to the next post-training run. Both are \emph{capital}---past tokens that lower the marginal cost or raise the marginal quality of future tokens.

\paragraph{Caches and memory as inventory.}
Let $S_t$ denote the stock of cached or memorized content (KV blocks \citep{kwon2023efficient}, retrieval embeddings \citep{lewis2020retrieval}, or agent notes \citep{park2023generative}). Its dynamics are
\begin{equation}
\label{eq:cap}
    S_{t+1} \;=\; (1-\delta_S)\,S_t \;+\; I_t,
\end{equation}
where $I_t$ is investment in new cache writes and $\delta_S$ captures distribution drift, schema change, and stale knowledge. The optimal-investment rule equates the marginal cost of writing to the discounted expected savings on future inference. Most production systems implement $S_t$ but rarely measure $\delta_S$, so cache hit rate is reported as an accounting metric rather than an economic one. Reusing a cached prefix when the new task value $V(x')$ differs from the original $V(x)$ is a quality externality; the correction is to track provenance and reuse only when expected reuse value clears an explicit threshold derived from Equation~\ref{eq:master}.

\paragraph{RL post-training as token investment.}
Reasoning-oriented post-training \citep{ouyang2022training,bai2022training,deepseek2025r1,jaech2024openai} consumes tokens that no end-user reads: rollouts, reward computations, KL-regularized updates \citep{schulman2017proximal,rafailov2023direct,ahmadian2024back}. These tokens are not consumption but \emph{investment} in future model capability. The right analogy is the neoclassical capital-accumulation model \citep{solow1956contribution}. With $A_t$ the model capability and $R_t, V_t, U_t$ tokens spent on rollouts, verification, and updates,
\begin{equation}
\label{eq:capacc}
    A_{t+1} \;=\; A_t \;+\; g(R_t, V_t, U_t) \;-\; \delta\, A_t,
\end{equation}
the optimal allocation equalizes marginal capability gain per token spent across modes: $\frac{\partial g/\partial R_t}{\kappa_R} = \frac{\partial g/\partial V_t}{\kappa_V} = \frac{\partial g/\partial U_t}{\kappa_U}$, where $\kappa_R, \kappa_V, \kappa_U$ are the per-token shadow prices introduced in Eq.~\ref{eq:bellman}. Equation~\ref{eq:capacc} embeds a Bellman problem; with discount $\beta \in (0,1)$ and per-token shadow prices $\kappa_{(\cdot)}$,
\begin{equation}
\label{eq:bellman}
    W(A_t) \;=\; \max_{R_t, V_t, U_t}\; \big\{\, \pi(A_t) - \kappa_R R_t - \kappa_V V_t - \kappa_U U_t + \beta\,\mathbb{E}\,W(A_{t+1}) \,\big\}.
\end{equation}
SFT, DPO \citep{rafailov2023direct}, and online RL \citep{schulman2017proximal,deepseek2025r1} are token-investment assets with different risk--return profiles: SFT is low-variance imitation, DPO is preference-bounded, online RL is high-variance exploration whose returns depend on verifier quality \citep{cobbe2021training,lightman2024lets}. Verifier tokens are risk capital---cutting them is identical to cutting risk capital in a financial firm: it lowers measured cost and raises tail risk \citep{ahmadian2024back}. The cost structure of online RL is itself shaped by the supply layer: speculative rollouts that share a tree-structured cache across trajectories \citep{chang2026srtacceleratingreinforcementlearning} lower $\kappa_R$ by amortizing prefix computation, which under Equation~\ref{eq:bellman} should shift the optimal mix toward more rollout tokens and away from purely imitation-based investment. In agentic post-training, the situation is further complicated because the same trace produces tool calls, plans, and final answers; cleanly separating those concerns at the pipeline level \citep{zhu2026opentinkerseparatingconcernsagentic} is what allows the planner to assign distinct shadow prices $\kappa_R, \kappa_V, \kappa_U$ to the components of an otherwise monolithic ``RL token.''

\paragraph{Portfolio frontier and closing the loop.}
SFT, DPO, and online RL form a portfolio of token-investment assets: SFT is low-variance, short-payback imitation; online RL is high-variance, longer-payback exploration; DPO sits between, and verifier tokens act as risk capital lowering the variance of every other asset's return. The Markowitz logic \citep{markowitz1952portfolio} predicts that the efficient frontier is a mix, not a corner---which is why ``all-RL'' or ``all-SFT'' pipelines typically underperform mixed schedules, and why aggressive verifier cuts tighten short-term budgets but blow up long-run learning curves. After this trace---now potentially training data---the same token has passed through all four layers, priced in dollars, risk, latency, and discounted future capability respectively. The four prices were never identical and were never visible to a single optimizer; they had to be reconciled by Equation~\ref{eq:master}. Agentic AI is one allocation problem, not four.

\paragraph{Position.}
Caches and RL pipelines should be reported with a depreciation rate, a hit-rate decomposition by $V(x)$, and a marginal-capability-per-investment-token estimate. Without these, ``cache hit rate'' and ``rollout volume'' are accounting metrics rather than economic ones.

\section{The Cost of Local Optimization}
\label{sec:failures}

We have followed one request through four layers and seen that each layer's mechanism is a different reading of Equation~\ref{eq:master}. We now turn from synthesis to diagnosis. The unified view sharpens what counts as a failure: a system fails not when it is slow or expensive in absolute terms, but when its allocation deviates predictably from Equation~\ref{eq:master}. The seven failure modes in Table~\ref{tab:failures} are not independent observations across heterogeneous systems; they are the corner cases of Equation~\ref{eq:master} when one of the four prices ($V$, $\Delta C_i$, $\lambda$, $\rho$) is held at zero or at infinity by a layer that does not see it.

\begin{table}[t]
  \caption{Marginal token allocation predicts a small set of recurring failure modes across heterogeneous LLM systems. Each row is a violation of Equation~\ref{eq:master} or Equation~\ref{eq:equim}.}
  \label{tab:failures}
  \centering
  \small
  \begin{tabular}{lll}
    \toprule
    Failure mode & Allocation violated & Where observed \\
    \midrule
    Over-routing       & Marginal $V\Delta Q_m < \Delta C_m$ for chosen $m$ & Frontier-default deployments \\
    Under-routing      & $V\Delta Q_m \gg \Delta C_m$ ignored & Cost-minimizing routers \\
    Over-delegation    & $\partial R/\partial a$ exceeds $V\,\partial p/\partial a$ & Auto-execute coding/email agents \\
    Under-verification & $V\Delta Q_v - \rho\Delta R_v$ positive but $T_v=0$ & Skip-the-tests pipelines \\
    Serving congestion & $\lambda \Delta L_i$ un-priced in $\Delta C_i$ & Flat-rate inference APIs \\
    Stale RL rollouts  & $\delta A_t$ exceeds $g(\cdot)$ at the margin & Long async PPO loops \\
    Cache misuse       & Reused KV with mismatched $V(x)$ & Naive prefix-cache reuse \\
    \bottomrule
  \end{tabular}
\end{table}

\paragraph{Why the same failure recurs.}
Heterogeneous teams---router authors \citep{chen2023frugalgpt,ong2024routellm}, agent authors \citep{yao2023react,wang2024voyager}, serving authors \citep{kwon2023efficient,zhong2024distserve}, RL authors \citep{deepseek2025r1}---repeatedly under-price the same quantity. The structural reason is that the four prices in Equation~\ref{eq:master} sit at different layers: $V$ is exposed to the user, $\Delta C_i$ to the operator, $\lambda$ to the SLA, and $\rho$ to the safety team. Locally rational decisions---``my router minimizes cost,'' ``my serving stack maximizes throughput,'' ``my agent maximizes success rate''---compose into globally irrational allocations. The prescription is not better local optimizers but a shared accounting object \citep{tirole1988theory}.

\paragraph{Equilibrium across tenants.}
In multi-tenant deployments the failures interact. A heavy-context tenant raises $\lambda$ for everyone via congestion; an aggressive autonomy tenant raises $\rho$ via reputational risk; a high-volume RL tenant raises $\Delta C$ on inference capacity. The right object is a competitive equilibrium in which shadow prices clear across tenants \citep{mascolell1995micro}, not a single-tenant optimization. Few production systems run such an equilibrium today; we view this as the next layer of the design problem.

\paragraph{Diagnosis vs. dashboard.}
A practical implication is that current dashboards measure the wrong things. ``Tokens per dollar'' is the average compute productivity; ``p95 latency'' is the supply-layer congestion summary; ``win rate'' is the demand-layer quality summary. None of them reads off Equation~\ref{eq:master}. A token-aware dashboard would instead report, per request, the realized vector $(V, \Delta C_i, \lambda \Delta L_i, \rho \Delta R_i)$ and the gap between realized and ex-post optimal allocation. This is harder to implement, but it is the only metric the framework treats as informative: every other dashboard captures a marginal slice and risks Goodhart's-law optimization at the layer that owns it.

\paragraph{Empirical predictions.}
The framework is falsifiable at the system-design level. Three predictions follow directly. First, holding $V$ fixed, raising the agent's verifier budget $T_v$ should monotonically reduce realized risk $R(a, t)$ until the marginal product of $T_v$ matches its marginal cost; agents whose verifier budget is below this point should reliably under-perform on high-$\rho$ tasks. Second, the same router that minimizes operator cost should display systematic regret on long-tail high-$V$ requests, identifiable from logs by the gap between achieved and ex-post optimal model. Third, multi-tenant serving stacks that flat-price tokens should observe quality regressions correlated with the volume of long-context traffic, even when none of the regressing tenants used long contexts themselves---the characteristic fingerprint of an unpriced congestion externality. Each of these predictions can be checked against existing production traces.

\section{Alternative Views}
\label{sec:alt}

\paragraph{Token economics is a metaphor, not a theory.}
A reasonable critic will argue that ``marginal,'' ``screening,'' and ``investment'' are loose analogies. The analogies are formal, not rhetorical: each layer reduces to a first-order condition (Equations~\ref{eq:master}, \ref{eq:routing}, \ref{eq:autonomy}, \ref{eq:serving}, \ref{eq:capacc}) testable on logs. The framework is falsifiable: a system that violates the relevant first-order condition should be Pareto-dominated by one that does not, and this can be checked empirically.

\paragraph{The right primitive is FLOPs, not tokens.}
Compute-optimal scaling work \citep{hoffmann2022training,kaplan2020scaling} argues for FLOPs as the natural budget. We agree FLOPs are correct for pre-training. For agentic systems, however, the binding constraints are increasingly latency, action risk, and verifier quality---not raw FLOPs. A FLOP spent on prefill, on a verifier, and on a tool call is economically distinct, and tokens (not FLOPs) preserve that distinction.

\paragraph{Optimization, not economics, is the right frame.}
A well-known alternative is to treat all of this as constrained optimization or RL: write down the reward and let gradient descent allocate. We do not disagree about the implementation; we argue that economics provides the \emph{specification}. Equilibrium concepts, screening, and externalities tell us \emph{which} reward to optimize and \emph{what counts as a market failure}. Without that specification, one is free to optimize the wrong objective extremely efficiently---a recurring pattern when token cost is minimized while risk-adjusted utility falls.

\paragraph{Centralized planners outperform marginal rules.}
A trainer could in principle solve a global plan over routing, agent policy, serving, and RL training jointly. A centralized planner is a valid algorithmic target, but it must still know which prices are being minimized against which constraints. Marginal allocation supplies that language and decomposes the joint problem into auditable subproblems.

\paragraph{This view will be obsolete when tokens are abolished, or tokens are mere billing artifacts.}
Two opposing critiques converge on the same point. Some argue that latent-space agents or continuous-action policies will make ``tokens'' an artifact; others argue that token billing has itself produced bad incentives (e.g., chain-of-thought becoming a billing strategy). The framework absorbs both: the load-bearing concept is \emph{marginal allocation}, not the token, and a billing artifact decoupled from $\Delta C_i$ in Equation~\ref{eq:master} is precisely the kind of Pigouvian distortion the framework is designed to diagnose.

\section{Discussion}
\label{sec:discussion}

\paragraph{Implications for system design.}
Five design and evaluation principles follow directly from Equation~\ref{eq:master}. \emph{Token-aware evaluation} should report the four prices ($V$, $\Delta C_i$, $\lambda$, $\rho$) and the realized allocation per request, not only aggregate accuracy and dollar cost. \emph{Risk-adjusted routing} should publish a regret bound against Equation~\ref{eq:routing} or an incentive-compatible menu (Eq.~\ref{eq:ic}), not a cost--quality scatter plot. \emph{Autonomy pricing} should make the action class explicit and price irreversible actions higher than reversible ones, in line with Equation~\ref{eq:autonomy}. \emph{Congestion-priced serving} should expose shadow prices for prefill, decode, and KV resources, so that upstream allocators can read them in real time and respond to the operator's binding constraints rather than to a flat per-token list price. \emph{RL token budgeting} should equalize marginal capability gain across rollouts, verifiers, and updates (Eq.~\ref{eq:bellman}) and depreciate stale rollouts at the rate $\delta$ implied by drift, not at the rate implied by an arbitrary epoch boundary. None of these principles requires new mathematics beyond Section~\ref{sec:theory}; what they require is a single, instrumented price vector visible to all four layers.

\paragraph{Limitations.}
We deliberately stop at a first-order condition; we make no claim that summing Equation~\ref{eq:master} across tasks yields a complete macroeconomic theory of AI. Three limitations deserve explicit acknowledgement. First, our prices treat compute, latency, and risk as commensurable in dollar units; this is a simplification that breaks down when physical or regulatory constraints are absolute (energy caps, data-residency rules) and require lexicographic rather than scalar treatment. Second, the framework assumes that $V(x)$ is at least partially observable; tasks whose value is realized only after long horizons (research-grade scientific reasoning, multi-month software engineering) are poorly captured by a one-step marginal rule and may require a multi-period extension. Third, our welfare-theorem argument (\S\ref{sec:theory}) presumes convexity of the production frontier and absence of strategic gaming on either side; LLM markets violate both at the seams, and the gap between the idealized equilibrium and the implementable mechanism remains open.

\section{Conclusion}
\label{sec:conclusion}

We have argued that agentic AI systems should be designed and evaluated as marginal token allocation economies. The argument is built on three load-bearing claims. First, four ostensibly separate layers---routing, agent policy, serving, and post-training---are vertical slices of a single allocation problem characterized by Equation~\ref{eq:master}, with prices that are formally Lagrange multipliers of the joint feasibility set. Second, recurring failures across the stack (over-routing, over-delegation, under-verification, congestion, stale rollouts, cache misuse) are corner cases of that equation when one of the four prices is mis-set, and they are predictable rather than incidental. Third, a Pareto-efficient allocation across the four layers requires only that the layers see a common, complete price vector---a condition that current production stacks systematically fail. The prescription is not centralization; it is shared price discovery. Returning to the developer with a failing test, the request is not a single completion but a chain of allocations: model tier, action authority, serving resources, and future training value. Today's systems price these decisions separately, producing silent downgrades, runaway autonomy, latency spikes, and noisy learning signals. The next generation of agentic AI systems will not be defined only by cheaper tokens or larger models, but by mechanisms that allocate marginal computation closest to the risk-adjusted equilibrium.

\bibliographystyle{plainnat}
\bibliography{references}

\newpage
\appendix

\section{Open Problems}

The framework leaves a focused set of open problems. (1) \emph{Estimation of $\Delta Q_i$ from logs} via causal inference / off-policy evaluation \citep{ouyang2022training}, with calibrated variance. (2) \emph{Risk pricing}: an empirical proxy for $\rho \Delta R_i$ that incorporates the Knightian component of Section~\ref{sec:theory}. (3) \emph{Mechanism-design routing}: do incentive-compatible menus (Eq.~\ref{eq:ic}) outperform silent routing under strategic users, and how should reasoning budgets be calibrated to per-task certainty signals \citep{fu2025efficientlyscalingllmreasoning}? (4) \emph{Internal shadow prices}: serving APIs that expose prefill, decode, and KV shadow prices upstream, building on schedulers that already learn request-level priorities \citep{fu2024efficientllmschedulinglearning}. (5) \emph{RL portfolios}: when SFT, DPO, and online RL---together with architectural variants such as speculative rollouts \citep{chang2026srtacceleratingreinforcementlearning} and concern-separated agentic pipelines \citep{zhu2026opentinkerseparatingconcernsagentic}---are treated as token-investment assets, what is the efficient frontier in the (variance, capability gain) plane? (6) \emph{Distributed equilibrium}: can the multi-tenant equilibrium of Section~\ref{sec:theory} be implemented as a clearing protocol, or must it be approximated by admission control plus priority queueing, and how should caches report depreciation $\delta_S$ in Equation~\ref{eq:cap}?

\section{Broader impact}
Treating agentic AI as a token economy makes \emph{who pays for what} explicit, which we view as a prerequisite for accountability. A user whose request is silently downgraded does not currently see the routing decision; a tenant whose latency degrades because of an unrelated long-context workload cannot identify the externality; a workforce whose tasks are delegated to an autonomous agent has no menu of oversight intensities to choose from. Instrumented prices make these decisions auditable, which is a public good. They are not, however, a substitute for governance: a mis-set $\rho$ on irreversible actions can still cause harm at speed, and an information-rent-extracting router can still be unfair even if it is welfare-maximizing in expectation. The framework should be read as a tool for diagnosis and design, not as a normative claim that markets settle every question. In particular, we are not arguing that decentralized token markets will spontaneously solve agentic-AI design; the history of computation markets \citep{coase1937nature,tirole1988theory} shows that decentralization without instrumented prices typically produces pathological equilibria, and current LLM markets---bundled pricing, opaque routing, unpriced congestion---are precisely such an environment. The argument is that agentic systems should be designed with the prices written down, so that internal optimization, external pricing, and human oversight are aligned to the same first-order condition.


\end{document}